\documentclass[conference]{IEEEtran}
\IEEEoverridecommandlockouts
\usepackage{cite}
\usepackage{amsmath,amssymb,amsfonts}
\usepackage{algorithmic}
\usepackage{graphicx}
\usepackage{textcomp}
\usepackage{verbatim}
\usepackage{bm}
\usepackage{tablefootnote}
\usepackage{threeparttable}
\usepackage{xcolor}
\usepackage{multirow}

\let\OLDthebibliography\thebibliography
\renewcommand\thebibliography[1]{
  \OLDthebibliography{#1}
  \setlength{\parskip}{0pt}
  \setlength{\itemsep}{0pt plus 0.3ex}
}
\def\BibTeX{{\rm B\kern-.05em{\sc i\kern-.025em b}\kern-.08em
    T\kern-.1667em\lower.7ex\hbox{E}\kern-.125emX}}
\begin{document}

\title{{TLRM:} Task-level Relation Module for GNN-based Few-Shot Learning
}


\author{\IEEEauthorblockN{Yurong Guo$^1$, Zhanyu Ma$^{1*}$, Xiaoxu Li$^2$, and Yuan Dong$^1$}
\IEEEauthorblockA{$^1$ Pattern Recognition and Intelligent System Lab.,\\ Beijing University of Posts and Telecommunications, Beijing, China}
\IEEEauthorblockA{$^2$ Lanzhou University of Technology, Lanzhou, China}
\thanks{$^*$ Corresponding author}
}

\maketitle

\begin{abstract}
Recently, graph neural networks (GNNs) have shown powerful ability to handle few-shot classification problem, which aims at classifying unseen samples when trained with limited labeled samples per class. GNN-based few-shot learning architectures mostly replace traditional metric with a learnable GNN. In the GNN, the nodes are set as the samples’ embedding, and the relationship between two connected nodes can be obtained by a network, the input of which is the difference of their embedding features. We consider this method of measuring relation of samples only models the sample-to-sample relation, while neglects the specificity of different tasks. That is, this method of measuring relation does not take the task-level information into account. To this end, we propose a new relation measure method, namely the \textbf{\textit{task-level relation module (TLRM)}}, to explicitly model the task-level relation of one sample to all the others. The proposed module captures the relation representations between nodes by considering the sample-to-task instead of sample-to-sample embedding features. We conducted extensive experiments on four benchmark datasets: mini-ImageNet, tiered-ImageNet, CUB-$200$-$2011$, and CIFAR-FS. Experimental results demonstrate that the proposed module is effective for GNN-based few-shot learning.
\end{abstract}

\vspace{-3mm}

\section{Introduction}
Deep Learning has been achieved great success in visual recognition tasks~\cite{vgg,resnet,9005389,9350209}, which depends on powerful model and amounts of labelled samples~\cite{imageNet}. However, humans can learn new concepts with little examples, or none at all. The gap motivated researchers to study few-shot learning and zero-shot learning.

The goal of few-shot learning is to classify unseen samples, given just a small number of labeled samples in each class. It has attracted considerable attention~\cite{Prototypical,Relation,GNN,siamese,Edge-Labeling,Transductive, CovaMNet,DN4,9293172,Xu_2021_CVPR,Zhang_2021_CVPR,zhang2021prototype}. One promising study is metric-based few-shot learning~\cite{Prototypical,Relation,GNN,siamese,Edge-Labeling,Transductive, CovaMNet,DN4}. Given just a query sample and a few labeled support samples, the embedding function extracts feature for all samples, and then a metric module measures distance between the query embedding and class embedding to give a recognition result. Recently, there have some studies of utilizing Graph Neural Networks (GNNs)~\cite{GNN,Edge-Labeling,Transductive,DPGN} to handle the few-shot classification task, which can be seen as a kind of metric learning method. 
In GNN-based few-shot learning model, all embedding features are connected to construct a graph. And each node is represented by the embedding feature of a sample. Then the graph classifies the unlabeled query by measuring the similarity between two samples.

\begin{figure}
	\centering
	\includegraphics[width=0.5\textwidth]{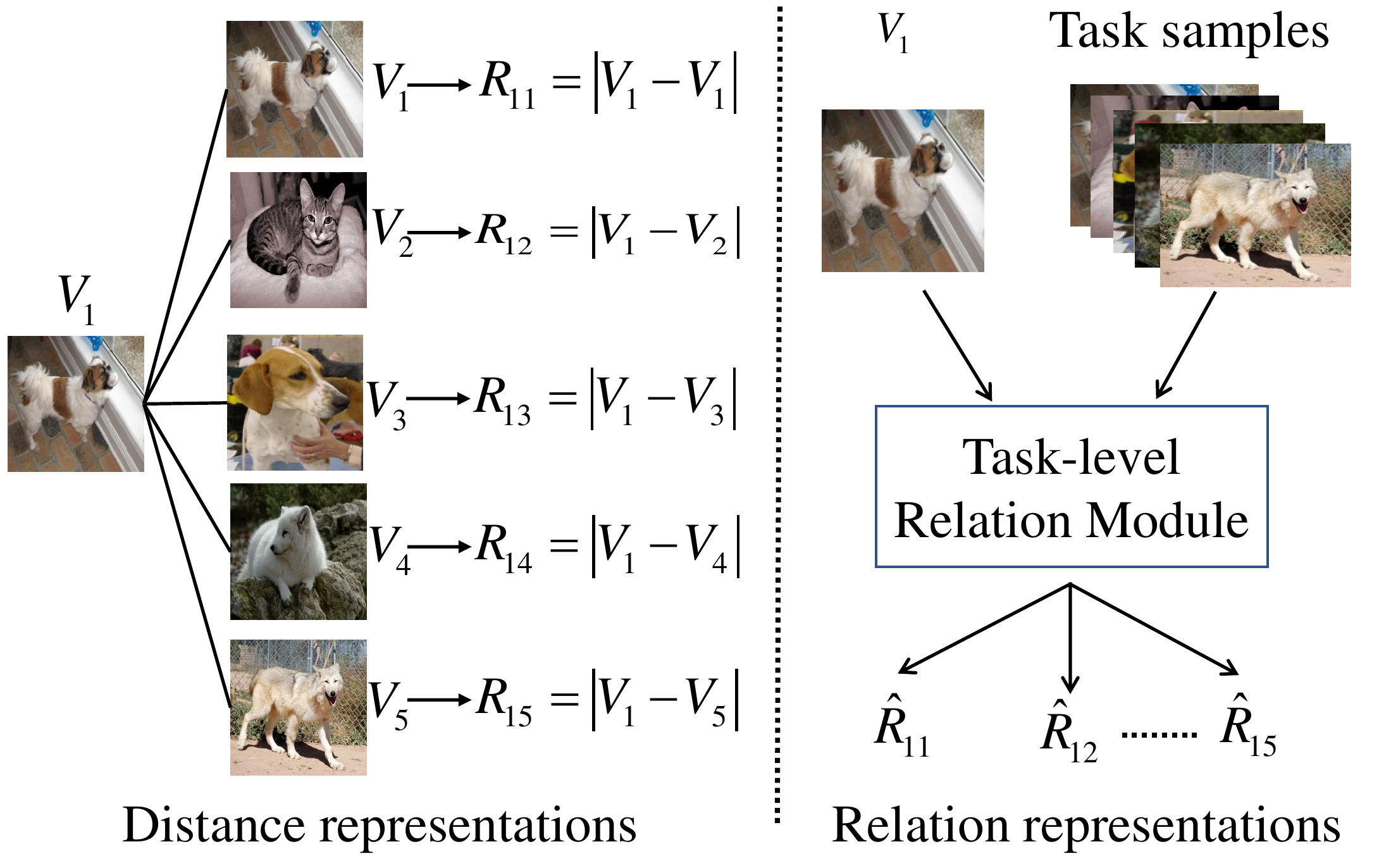}
	\caption{The left panel shows a general framework of previous metric approaches based on GNN. 
	 The right panel briefly illustrates our approach.}
	\label{fig1}
	\vspace{-5mm}
\end{figure}

\begin{figure*}
	\centering
	\includegraphics[width=0.98\textwidth]{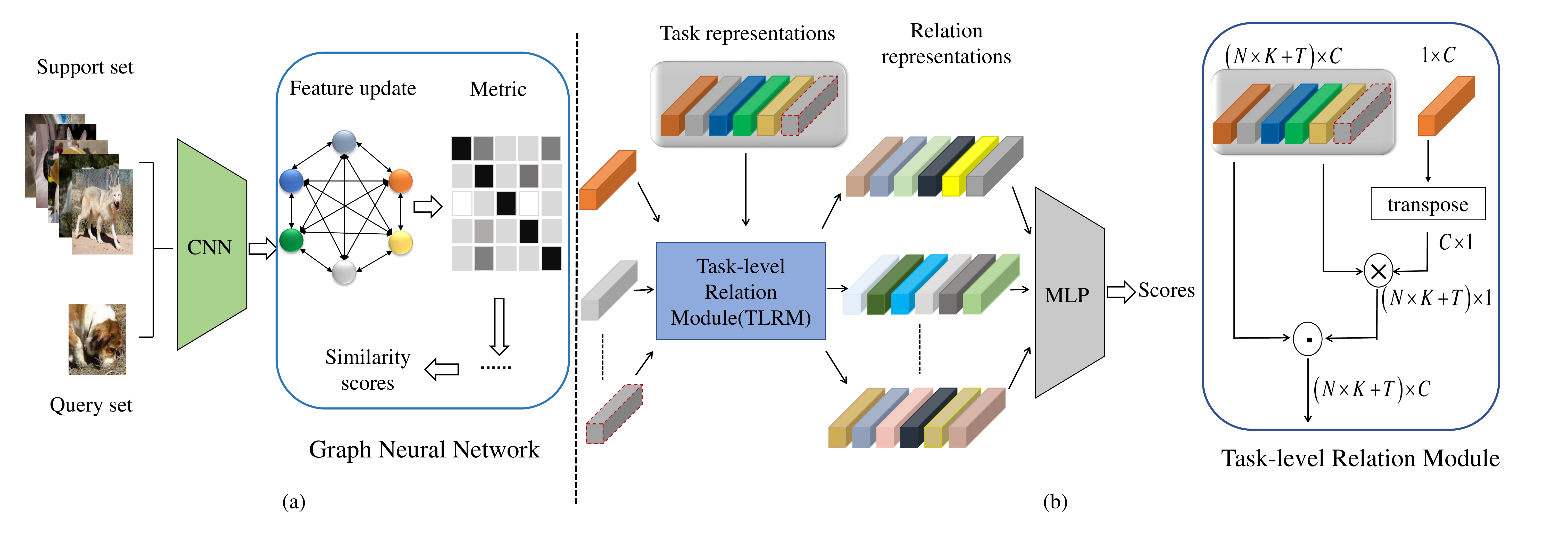}
	\caption{(a) shows the GNN-based few-shot model. And in (b), the left panel shows a general framework of our approach used for calculating the similarity scores, and the right panel shows the task-level relation module.}
	\label{fig3}
	\vspace{-5mm}
\end{figure*}

Even though GNN-based model have made significant advance in few-shot classification, they do suffer from distinct limitation.
In the metric module of GNN-based methods, relation representation for a pair samples is obtained by calculating the absolute difference~\cite{GNN,Edge-Labeling,Transductive,DPGN}. It only considers the corresponding embedding features of the samples. Intuitively, pair-wise relationship is not only dependent on the distance between corresponding embedding features, but also related to all embedding features in a task. As shown in the left panel of Figure \ref{fig1}, there is no significant difference between the target sample and all other samples in the task. The distance representation between two samples neglects the specificity of the task and lacks  discrimination. This will cause the problem that the similarity scores are not significantly different, so that the category of the target sample is not clear. To deal with the key challenge of how to learn relation representations with distinctive information, we propose a sample-to-task metric module, as shown in the right panel of Figure \ref{fig1}, which adopts a meta learning strategy to learn the relation representations.
The main contributions of this paper are summarized as follows:

\begin{itemize}
\item	We propose an \textbf{\textit{task-level relation module (TLRM)}}. The proposed TLRM utilizes the attention mechanism to learn task-specific relation representations for each task. 
	
\item  The comprehensive experimental  results  on four benchmark datasets show that our proposed module is effective for GNN-based few-shot model. In addition, the results of semi-supervised few-shot classification and visualization of similarity scores are provided to further evaluate our module.
\end{itemize}

\vspace{-1mm}
\section{Related work}
\label{sec:Related work}

\textbf{Meta Learning in Few-shot Learning:} 
Meta Learning  framework is an effective study for few-shot learning, which  mainly focuses on how to learn and utilize meta-level knowledge to adapt to new tasks quickly and well. One of the excellent studies is model-agnostic meta-learning (MAML)~\cite{MAML}. The MAML learned initialization parameters by cross-task training strategy such that the base learner  can rapidly generalize new tasks using a few support samples. 
Subsequently, many MAML variants~\cite{task_1, latent_Meta, caml, MAML-lstm, MetaNet, LGM-Net} have been developed.

\textbf{Metric Learning in Few-shot Learning:} 
On the metric learning side, most of algorithms consist of embedding function extracting features for instances and metric function for measuring sample between the query embedding and class embedding. Koch et al.~\cite{siamese} used siamese network to compute the pair-wise distance between samples. Prototypical networks~\cite{Prototypical} firstly built a prototype representation of each class  and measured the samples between the query embedding and class’s prototype by using euclidean distance.
Matching network~\cite{Matching-Networks} used a neural network with external memories to map samples to embedding features, which considers full context in a task. 
TADAM~\cite{TADAM} introduced a metric scaling factor to optimize the similarity metric of prototypical nets.
Zheng et al.~\cite{Principal_characteristic} believed that the average prototype ignores the different importance of different support samples and proposed principal characteristic nets.

Fixed metric methods will restrict the embedding function to produce discriminative representations. Sung et al.~\cite{Relation} introduced relation network (RN) for few-shot learning. The relation network learns to learn a deep distance metric by a neural network. However, due to the inherent local connectivity of CNN, the RN can be sensitive to the spatial position relationship of semantic objects in two compared images. To address this problem, Wu et al.~\cite{PARN} introduced a deformable feature extractor (DFE) to extract more efficient features, and designed a dual correlation attention mechanism (DCA) to deal with its inherent local connectivity. 
Hou et al.~\cite{Cross-Attention} proposed a cross attention network for few-shot classification, which is designed to model the semantic relevance between class and query features. 


\textbf{GNN-based methods in Few-shot Learning:}
Recently, most approaches are proposed to exploit GNN in the field of few-shot learning task. Specifically, Garcia et al.~\cite{GNN} first utilized GNN to solve few-shot learning problem, where all embedding features  extracting by a convolutional neural network are densely connected.  Liu et al.~\cite{Transductive} proposed a transductive propagation network (TPN). The TPN utilizes the entire query set for transductive inference. To further exploit intra-cluster similarity and inter-cluster dissimilarity, kim et al.~\cite{Edge-Labeling} proposed an edge-labeling graph neural network. Then in order to explicitly model the distribution-level relation,  Yang et al.~\cite{DPGN} proposed distribution propagation graph network (DPGN).

In the existing GNN-based few-shot learning methods, pair-wise distance representations are absolute difference of the embedding features. However, when the classes in the task are  similar, it will lead to the problem of insufficient discrimination in metric representations. So, in this paper, we focus on learning distinctive relation information through an task-level relation module. 

\section{The Proposed Method}
\label{sec:Method}
\subsection{GNN-based few-shot learning}
\label{ssec:GNN-based few-shot}

\begin{table*}[ht]
	\centering
	\caption{$5$-way $1$-shot classification accuracy on four benchmark datasets: mini-ImageNet, tiered-ImageNet, CUB-$200$-$2011$, and CIFAR-FS}
	
    \vspace{-1mm}
	\begin{tabular}{c|c|c|c|c|c}
		\hline
		Model                & Trans. & mini-ImageNet & tiered-ImageNet & CUB-200-2011        & CIFAR-FS   \\ \hline
		EGNN (CVPR 19)               & No     & $52.86\pm0.42$      & $57.09\pm0.42$         & $64.82\pm  0.41$ & $65.51\pm0.43$ \\ 
		EGNN + TLRM          & No     & $53.65\pm0.43$    & $57.40\pm0.42$      & $65.07\pm0.41$ & $65.00\pm0.42$ \\ \hline \hline
		TPN (ICLR 18)              & Yes    & $59.46$        & $-$          & $-$ & $-$ \\
		EGNN                & Yes    & $58.94\pm0.51$        & $62.37\pm0.51$          & $73.18\pm0.51$ & $72.20\pm0.49$ \\
		DPGN (CVPR 20)             & Yes    & $66.41\pm0.51$    & $71.86\pm0.50$      & $75.25\pm0.46$ & $75.83\pm0.47$ \\ \hline
		EGNN + TLRM          & Yes    & $60.79\pm0.51$    & $64.52\pm0.51$      & $75.02\pm0.46$ & $73.42\pm0.50$ \\ 
		DPGN + TLRM          & Yes    & \bm{$66.97\pm0.53$}    & \bm{$72.24\pm0.50$}      & \bm{$77.53\pm0.46$} & \bm{$77.05\pm0.46$}  \\ 
		\hline
	\end{tabular}
   \vspace{-3mm}
      	\label{table1}
\end{table*}

\begin{table*}[]

	\centering
	    
	\caption{$5$-way $5$-shot classification accuracy on four benchmark datasets: mini-ImageNet, tiered-ImageNet, CUB-$200$-$2011$, and CIFAR-FS}
	\vspace{-1mm}
    \begin{threeparttable}
	\begin{tabular}{c|c|c|c|c|c}
		
		\hline
		Model       & Trans. & mini-ImageNet & tiered-ImageNet & CUB-200-2011       & CIFAR-FS   \\ \hline
		EGNN       & No     & $68.20\pm0.41$        & $71.13\pm0.39$          & $80.05\pm0.36$ & $76.95\pm0.37$ \\ 
		EGNN + TLRM & No     & $68.72\pm0.40 $    & $72.39\pm0.39 $     & $81.03\pm0.36$ & $77.78\pm0.37$ \\ \hline \hline
		TPN                & Yes    & $75.65$        & $-$          & $-$ & $-$ \\
		EGNN       & Yes    &$ 75.71\pm0.46 $      & $81.04\pm0.43$          & $87.68\pm0.38 $& $86.13\pm0.41 $\\  
		DPGN       & Yes    & $82.04\pm0.45 $   & $82.70\pm0.43 $     & $87.72\pm0.36$ & $87.85\pm0.38$ \\ \hline
		EGNN + TLRM & Yes    & $76.18\pm0.45 $   & $81.47\pm0.43$      &$ 88.00\pm0.36$ & $85.70\pm0.39 $\\ 
		DPGN + TLRM & Yes    & \bm{$82.58\pm0.45$}    & \bm{$83.31\pm0.44$}      & \bm{$90.39\pm0.34$} & \bm{$89.15\pm0.37$} \\ 
		\hline 

      	\end{tabular}

	\label{table2}

	      \begin{tablenotes}
        \footnotesize
        \item[ * ]  ``No'' means non-transductive method, and ``Yes'' means transductive method.
      \end{tablenotes}
      \end{threeparttable}
      \vspace{-5mm}
\end{table*}

As shown in Figure \ref{fig3} (a), GNN-based few-shot model usually consists of a CNN for extracting features and a GNN for propagating labels from labeled nodes to unlabeled according to similarity scores between nodes.  
In the training and testing process, GNN-based few-shot model usually adopts the episodic mechanism, in which each episode (task) consists of  the support set $S$ and the query set $Q$. And the support set contains $N*K$ labeled support samples and the query set contains  $T$  unseen samples in a $N$-way $K$-shot problem.

Generally, the CNN $g(\cdot)$ as backbone of extracting features has two different types  $1)$ the $4$-layer convolution network (ConvNet)~\cite{Transductive,Edge-Labeling} and $2)$ the $12$-layer residual network (ResNet-$12$) used in~\cite{DPGN}. The GNN consists of $L$ layers to process the graph. Let $V =  \left\{  V_1, V_2,...,V_{N\times K+T}\right\} $ be embedding features for all nodes extracted by the CNN,
$ R_{ij}$ be relation representations between nodes, and  $ s_{ij}$ be similarity score between node $i$ and $j$.  Given $V^{L-1}$ and $s^{L-1}$ from the layer $L-1$, node feature update is firstly conducted by a neighborhood aggregation procedure. And  node $i$ is updated as
\begin{equation}
V_i^L = f_v\left( \sum_{j=1}^{N\times K +T} V_j^{L-1}s_{ij}^{L-1} \right),
\end{equation}
where $ f_v$ is the feature (node) transformation network.
Then, the relation representation is obtained by calculating the absolute difference between two vector nodes. 	It can be denoted as
\begin{equation}
\begin{split}
R_{ij} &= \left| V_i^L - V_j^L \right| = \begin{matrix} \sum_{k=1}^C \left| V_{ik}^L - V_{jk}^L \right|\end{matrix} .
\end{split}
\end{equation}
Finally, the relation representation  $ R_{ij}$ is input into a Multilayer Perceptron (MLP) to capture the similarity scores between nodes
\begin{equation}
\begin{split}
s_{ij}&= f_s\left( R_{ij} \right)\\
& =  f_s\left( \begin{matrix} \sum_{k=1}^C \left| V_{ik}^L - V_{jk}^L \right|\end{matrix} \right)\\
&= \sigma \left ( \begin{matrix} \sum_{k=1}^C \omega_k\left| V_{ik}^L - V_{jk}^L \right|\end{matrix} \right).
\end{split}
\end{equation}
Where $ f_s$ is  transformation network. The goal of GNN-based few-shot learning is to learn function $g$, $f_v$ and $f_s$ to classify query sample $x_{query}$ by $\hat{y}_{query} = f_s(f_v (g(x_{query};D_{support})))\in{(0,1)}^N$.
Note that the relationship is obtained by measuring the distance between two corresponding node, which is node-to-node and task-agnostic. 

\subsection{Task-level Relation Module}

In this paper, attention mechanism is employed to transform sample embedding to relation representations with consideration to task-specific embedding. Note that the relation representation is task-specific and not only the distance between nodes. We denote it as Task-level Relation Module (TLRM). The proposed TLRM can avoid direct comparison relative relationship irrelevant local representations. 
As shown in Figure \ref{fig3} (b), given the feature representations  $V\in R^{(N\times K+T) \times C}$, the relation representations can be obtained. The implementation details are performed as follows.

For node $i$, the attention value between the target embedding and all other samples in the task can be obtained by adopting method commonly used in the attention mechanism. The attention value is performed as follows
\begin{equation}
a\left(V_i, V_j \right) = \cfrac{\text{exp}{(e_{ij}})}{\begin{matrix} \sum_{k=1}^{N\times K +T} \end{matrix} \text{exp}{(e_{ik}})},
\end{equation}
Where $a\in R^{\left(N\times K +T\right)\times \left(N\times K +T\right)} $, which represents the similarity between nodes comparing to all other embedding in the task. $e_{ij}$ reflects the matching degree of node $i$ to node $j$. When the degree is higher, $a_{ij}$ is bigger. The matching degree $e_{ij}$  is performed as follows
\begin{equation}
 e_{ij} = s\left(V_j, V^T_i\right)/{\sqrt{C}},
\end{equation}
Where, first, the feature representation $V_i\in R^{1\times C}$ of target sample is reshaped to $V_i\in R^{C\times 1}$ through a transpose operation and 
 $ s\left(V_i, V^T_j\right) $ is the  vector multiplication operation. And then, $a_{ij}$ is used to encode $V_j$ and the relation representations can be obtained, which can be denoted as
\begin{equation}
\widehat{R}_{ij} = a\left(V_i, V_j \right)\odot V_j,
\end{equation}

The relation representation $\widehat{R}_{ij}$ models the relation representation between node $i$ and $j$, which is a task-level relation representation  of  sample $i$ to $j$ comparing to all the other samples.  Afterwards,  $\widehat{R}_{ij}$ is fed to an MLP to capture the relation score for performing further classification
\begin{equation}
s_{ij} = \text{MLP}(\widehat{R}_{ij}).
\end{equation}

\section{Experiments and Discussions}
\label{sec:Experiments}
\subsection{Datasets and setups}

To evaluate our module, we select two GNN-based few-shot models: EGNN and DPGN,  and  four standard few-shot learning benchmarks: mini-ImageNet~\cite{Matching-Networks}, tiered-ImageNet~\cite{tiered}, CUB-$200$-$2011$~\cite{WahCUB_200_2011} and CIFAR-FS~\cite{cifar}. 

For the sake of fairness, all experiments  employed the same setups as EGNN and DPGN.  EGNN used ConvNet, and DPGN used ResNet-$12$ for extracting  features. In training process, the Adam optimizer was used in all experiments with the initial learning rate $10^{-3}$. And the learning rate was decayed  by $0.1$ per $15000$ iterations. The weight decay was set as $10^{-5}$.
For all datasets, $5$-way $1$-shot and $5$-way $5$-shot experiments were conducted. We randomly sampled $10,000$ tasks and then reported the mean accuracy along with its $95\%$ confidence interval.

\begin{table}[]
	\centering
	\caption{Semi-supervised few-shot classification accuracy on mini-ImageNet. The results are tested in transductive learning.}
	\resizebox{\linewidth}{!}{
		\begin{tabular}{c|c|c|c}
			\hline
			Model       & $20\%$       & $40\%$       & $60\%$       \\ \hline
			EGNN       & $63.91\pm0.42$ & $65.84\pm0.41$ & $68.06\pm0.44$ \\ 
			\bf{EGNN + TLRM} & $\bf{66.37\pm0.41}$ & $\bf{66.82\pm0.42}$ & $\bf{69.08\pm0.42}$ \\ \hline
			DPGN       & 
			$74.16\pm0.44$ & $81.23\pm0.44$ &$ 80.84\pm0.46 $\\ 
			\bf{DPGN + TLRM} & $\bf{80.99\pm0.44}$ & $\bf{81.70\pm0.45}$ & $\bf{80.94\pm0.45}$ \\ \hline
	\end{tabular}}
	\label{table3}
	\vspace{-4mm}
\end{table}

\begin{table}[]
    \centering
    \caption{Effect of adding the proposed TLRM to different layer of EGNN. The results on $5$-way $1$-shot are reported.}
    \begin{tabular}{c|c|c}

        \cline{1-3}
        \multicolumn{2}{c|}{Model}            & mini-ImageNet   \\ \cline{1-3} 
        \multicolumn{2}{c|}{EGNN}             & $52.86\pm0.42$          \\ \cline{1-3}
        \multirow{4}{*}{EGNN+TLRM} & $L=1$       & $53.11\pm0.43$            \\ 
                                   & $L=2$       & $53.29\pm0.44$            \\ 
                                   & $L=3$       &    $53.42\pm0.42$              \\ 
                               &$ \bf{L=1 ~\&~2 ~\&~3}$ & $\bf{53.65\pm0.43}$          \\ \cline{1-3}
    \end{tabular}
    \label{table4}
    \vspace{-4mm}
\end{table}

\begin{figure}
	\centering
	\includegraphics[width=0.4\textwidth]{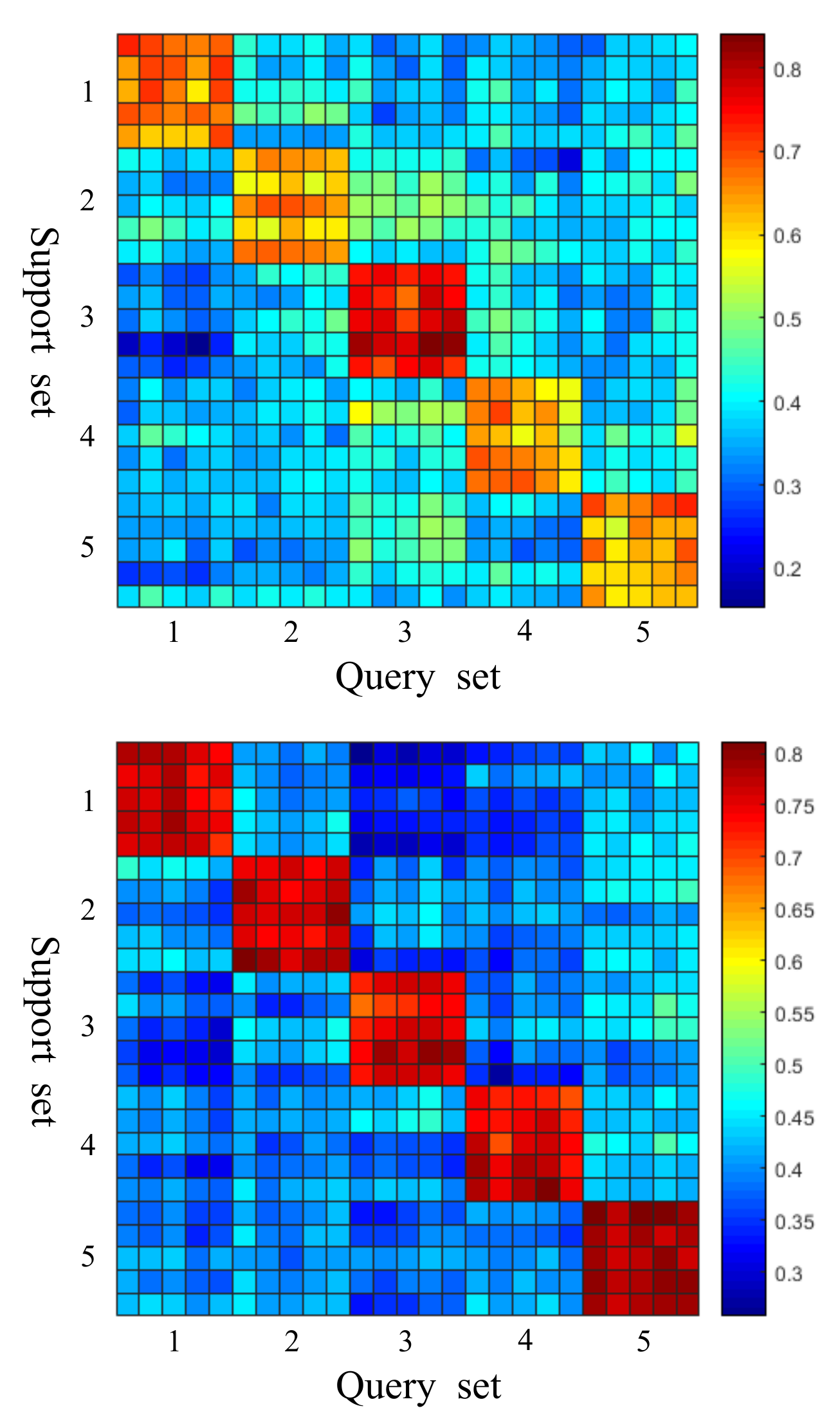}
	\vspace{-4mm}
	\caption{Visualization of similarity scores obtained by EGNN (top) and EGNN with our module (bottom).}
	\vspace{-5mm}
	\label{fig4}
\end{figure}


\vspace{-2mm}
\subsection{Results and discussions for few-shot classification}

Experimental results for $5$-way $1$-shot and $5$-way $5$-shot  classification are shown in Table \ref{table1} and Table \ref{table2}.  We can see that EGNN or DPGN with our TLRM have higher accuracy than the ones without TLRM on mini-ImageNet, tiered-ImageNet, and CUB-$200$-$2011$.
Meanwhile, partial experimental results on the CIFAR-FS dataset dropped slightly, the reason of which might lie in the categories in the CIFAR-FS dataset are highly distinguishable. In addition, the CUB-$200$-$2011$ dataset is the most widely used benchmark for fine-grained image classification, which has significant intra-class variance and inter-class similarity.  Fine-grained image task is more challenging in few-shot learning.  Clearly, the improvement on the CUB-$200$-$2011$ dataset is  significant in Table \ref{table1} and Table \ref{table2}, which shows that the relationship representation obtained by our module is more discriminative than previous method for tasks with high similarity. And overall, our method is simple and effective.

Semi-supervised experiments were conducted in $5$-way $5$-shot setting on mini-ImageNet with two backbones, in which the support samples are only partially labeled. The results are presented in Table \ref{table3}. Notably, the EGNN and DPGN with our TLRM outperforms the previous backbones especially when the labeled samples portion was decreased.

\subsection{Ablation studies}
In order to investigate the effect of our proposed TLRM on different layer of GNN,  ablation studies were conducted  with $L=1$, $L=2$, and $L=3$ on mini-ImageNet with EGNN backbone. It can be observed from Table \ref{table4} that the proposed TLRM plays a significant role in each layer of EGNN.

\subsection{Visualization of similarity scores} 
 For further analysis, Figure \ref{fig4} shows similarity scores in the last layer of EGNN. The similarity scores are the average of 10000 tasks in setting of 5-way 5-shot and 5 queries for each class. The 25 samples in vertical axis are support set, and 25 samples in horizontal axis are query set. Notably, EGNN  with our module not only contributes to predicting more accurately but also reduces the similarity score between samples in different classes and increases the similarity score between samples in the same classes. 
\vspace{-2mm}
\section{Conclusions}
\label{sec:Conclution}
In this paper, we propose an task-level relation module to capture the relation representations by employing all the embedding features in a single task. By considering all the samples in the task, our method can hold discriminative relation features for each node pair. Experimental results demonstrate that it improves the performance of recently proposed GNN-based methods on four benchmark datasets: mini-ImageNet, tiered-ImageNet, CUB-$200$-$2011$, and CIFAR-FS.


\bibliographystyle{IEEEbib}
\bibliography{conference_101719}

\end{document}